\documentclass[conference]{IEEEtran}
\IEEEoverridecommandlockouts

\usepackage{cite}
\usepackage{amsmath,amssymb,amsfonts}
\usepackage{algorithmic}
\usepackage{algorithm}
\usepackage{balance}
\usepackage{graphicx}
\usepackage{textcomp}
\usepackage{xcolor}
\def\BibTeX{{\rm B\kern-.05em{\sc i\kern-.025em b}\kern-.08em
    T\kern-.1667em\lower.7ex\hbox{E}\kern-.125emX}}
\begin{document}

\title{Efficient Transceiver Design for Aerial Image Transmission and Large-scale Scene Reconstruction
\thanks{This work is supported in part by the State Key Laboratory of Internet of Things for Smart City (University of Macau) Open Research Project under Grant SKL-IoTSC(UM)/ORP04/2026, and in part by the Project of Tsinghua University-Toyota Joint Research Center for AI Technology of Automated Vehicle under Grant TTAD-2024-08-2.}
}

\author{
    \IEEEauthorblockN{Zeyi Ren\IEEEauthorrefmark{1}, Jialin Dong\IEEEauthorrefmark{1}, Wei Zuo\IEEEauthorrefmark{2}, Yikun Wang\IEEEauthorrefmark{2}, Bingyang Cheng\IEEEauthorrefmark{2}, Sheng Zhou\IEEEauthorrefmark{1}, and Zhisheng Niu\IEEEauthorrefmark{1}}
    
    \IEEEauthorblockA{\IEEEauthorrefmark{1}Department of Electronic Engineering, Tsinghua University, Beijing, China}

    \IEEEauthorblockA{\IEEEauthorrefmark{2}Department of Electrical and Computer Engineering, The University of Hong Kong, Hong Kong}


%
    \IEEEauthorblockA{\fontsize{9pt}{12pt}\selectfont Emails: zeyiren0827@outlook.com, \{zuowei, ykwang, bycheng\}@eee.hku.hk, \{djl23@mails., sheng.zhou@, niuzhs@\}tsinghua.edu.cn}
}

\maketitle

\begin{abstract}
Large-scale three-dimensional (3D) scene reconstruction in low-altitude intelligent networks (LAIN) demands highly efficient wireless image transmission. However, existing schemes struggle to balance severe pilot overhead with the transmission accuracy required to maintain reconstruction fidelity. To strike a balance between efficiency and reliability, this paper proposes a novel deep learning-based end-to-end (E2E) transceiver design that integrates 3D Gaussian Splatting (3DGS) directly into the training process. By jointly optimizing the communication modules via the combined 3DGS rendering loss, our approach explicitly improves scene recovery quality. Furthermore, this task-driven framework enables the use of a sparse pilot scheme, significantly reducing transmission overhead while maintaining robust image recovery under low-altitude channel conditions. Extensive experiments on real-world aerial image datasets demonstrate that the proposed E2E design significantly outperforms existing baselines, delivering superior transmission performance and accurate 3D scene reconstructions.
\end{abstract}


\section{Introduction}
Within the emerging paradigm of low-altitude intelligent networks (LAIN)~\cite{LAIN_zhou}, unmanned aerial vehicles (UAVs) are playing an increasingly pivotal role in tasks such as autonomous exploration~\cite{auto_exploration}, infrastructure inspection~\cite{auto_inspection,icra_inspection}, and search-and-rescue~\cite{boyu_search}. These advanced applications are often required to perform large-scale three-dimensional (3D) scene reconstruction for comprehensive environmental perception~\cite{xinlei2025,gaofei}. However, the physical realization and computational efficiency of such high-fidelity scene reconstruction are governed by the efficient and reliable wireless transmission of massive aerial image data~\cite{STT-GS}. Therefore, developing an efficient and robust image transmission mechanism is a critical imperative for downstream reconstruction tasks in low-altitude scenarios.

However, achieving efficient transmission remains a challenge for traditional wireless transceivers. Conventional systems rely heavily on the insertion of pilot sequences to perform accurate channel estimation. Yet, in the highly dynamic channels typical of UAV tasks, tracking rapid fading requires massive pilot overhead~\cite{huawei}. This excessive overhead severely occupies the limited resources that should be dedicated to actual image payloads, fundamentally degrading the overall transmission efficiency.

\begin{figure} [t]
	\centering 
		 \includegraphics[width=0.48\textwidth]{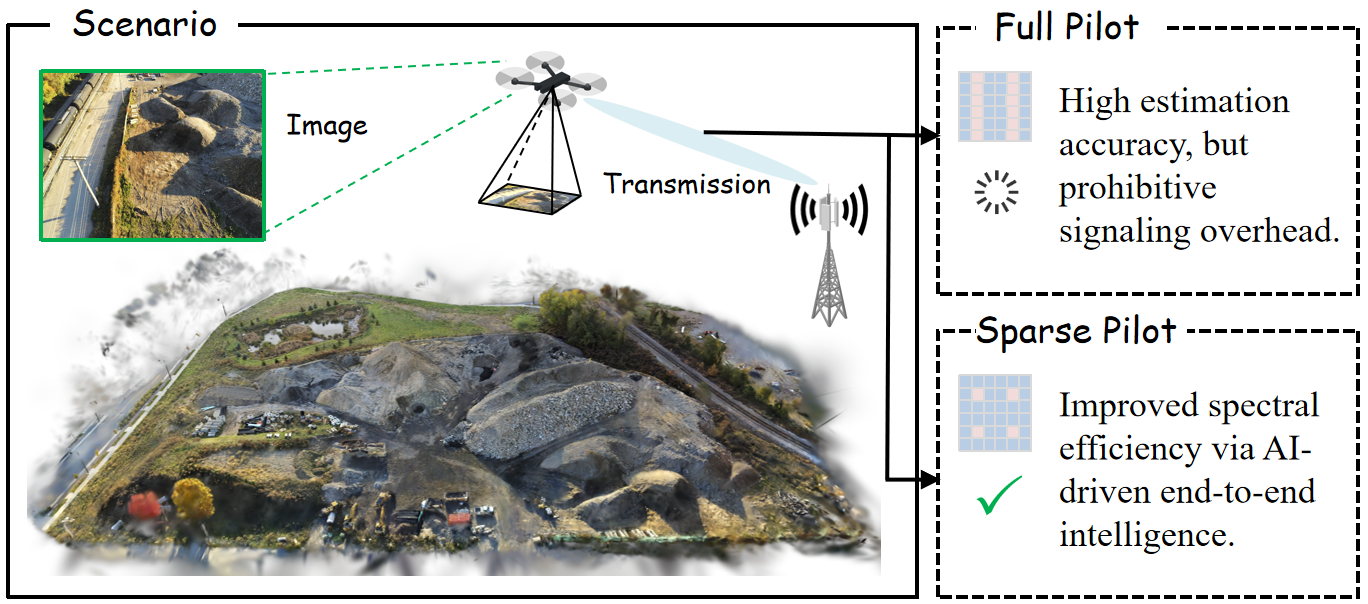} 
  	\caption{Illustration of the low-altitude intelligent network imagery scenario. Traditional full pilot schemes suffer from prohibitive signaling overhead, whereas the proposed AI-driven sparse pilot transmission improves spectral efficiency for downstream 3D scene reconstruction.}\vspace{-0.2in}  \label{Fig1}
\end{figure}

To alleviate this overhead, emerging pilotless transceivers maximize spectral efficiency but face severe reliability issues~\cite{Nvidia,aibo,lengyang,lizhong}. Without explicit pilot signals, adapting to fast-varying low-altitude channels relies entirely on end-to-end joint training. However, this process is fundamentally bottlenecked by extreme parameter asymmetry between the lightweight transmitter and the highly parameterized receiver. Consequently, erratic gradient propagation prevents learning robust constellations, ultimately compromising the downstream 3D scene reconstruction.

From the above arguments, a critical dilemma emerges: traditional massive pilots guarantee reliability at the cost of efficiency, whereas pilotless schemes maximize efficiency but fail to provide the robustness required for accurate reconstruction. To strike a balance between transmission efficiency and image fidelity, this paper proposes a novel deep learning-based end-to-end (E2E) transceiver design incorporating a sparse pilot scheme. Specifically, instead of using dense grids, this paper introduces a method that multiplexing a minimal number of pilot symbols into the data stream to serve as essential channel anchors~\cite{bin}. At the transmitter, a Transformer-based modulator is developed to learn channel-resilient constellation mappings. Concurrently, a ResNet-driven Deep-Rx network~\cite{Nvidia} is deployed at the receiver to jointly execute channel estimation and signal recovery based on these sparse anchors. By jointly optimizing the transmitter and receiver networks, our framework significantly reduces pilot overhead while maintaining robust image recovery under complex low-altitude channel conditions.

Additionally, conventional image transmission schemes optimize solely for pixel-level fidelity (e.g., mean squared error)~\cite{Qin_image}, neglecting the impact of different image regions on downstream 3D geometry~\cite{shuai_icassp}. To bridge the gap between physical-layer transmission and scene reconstruction, we directly integrate 3D Gaussian Splatting (3DGS)~\cite{kerbl3Dgaussians,shuai_gs} into the E2E training loop. By formulating a task-oriented loss function that jointly evaluates novel view synthesis quality and signal recovery, the transceiver implicitly learns to prioritize the transmission of geometric and photometric features critical for 3D reconstruction. This rendering-aware method ensures that under a restricted sparse pilot overhead, wireless resources are strategically allocated to maximize 3D scene fidelity.

Extensive evaluations on realistic aerial datasets demonstrate the dual advantages of the proposed E2E transceiver design. Compared to the ideal full-pilot transmission, our method achieves substantially higher efficiency while maintaining a nearly identical block error rate (BLER). Moreover, it demonstrates superior performance over conventional architectures and deep learning baselines in both transmission reliability and downstream 3D scene reconstruction quality.

\section{System Model}
Consider a downlink massive multiple-input multiple-output (MIMO) communication system where a single terrestrial base station (BS) serves an UAV. The BS is equipped with a uniform planar array (UPA) consisting of $N_{\mathrm{BS}}$ antennas and the UAV is equipped with a uniform linear array (ULA) of $N_{\mathrm{UAV}}$ antennas. 

\subsection{Spatial Configuration}
We adopt a 3D Cartesian coordinate system. The BS is located at a fixed position $\mathbf{p}_{\mathrm{BS}} = (0, 0, h_{\mathrm{BS}})$, where $h_{\mathrm{BS}}$ denotes the antenna height. The UAV is located at $\mathbf{p}(t) = (x(t), y(t), z(t))$ with a velocity vector $\mathbf{v}(t) \in \mathbb{R}^{3 \times 1}$ at time instant $t$. Due to the high altitude of UAVs, the communication links are characterized by high probabilities of Line-of-Sight (LoS) propagation and significant Doppler shifts caused by UAV mobility.

The horizontal distance between the BS and UAV is given by: $r = \sqrt{x^2 + y^2}$.
And the Euclidean distance is: $d = \sqrt{r^2 + (h_{\text{BS}} - z)^2}$.
The elevation angle from the BS to UAV plays a crucial role in UAV channel characteristics, which can be expressed as
\begin{equation}
    \theta = \arcsin\left(\frac{|h_{\text{BS}} - z|}{r}\right) \in \left[0, \frac{\pi}{2}\right],
\end{equation}
and the azimuth angle is modeled as
\begin{equation}
    \phi = \arctan\left(\frac{y}{x}\right) .
\end{equation}

\subsection{3D Non-Stationary UAV Channel Model}
Unlike traditional terrestrial channels, the air-to-ground (A2G) channel exhibits strong LoS components and 3D scattering characteristics. We adopt a geometry-based stochastic model (GBSM) extended from the 3GPP specifications~\cite{3gpp_tr_38901}. The uplink channel matrix $\mathbf{H} \in \mathbb{C}^{N_{\mathrm{BS}} \times N_{\mathrm{UAV}}}$ from the UAV to the BS can be modeled as a superposition of a deterministic LoS component and a stochastic Non-LoS (NLoS) component:
\begin{equation}
    \mathbf{H} = \sqrt{\beta} \left( \sqrt{\frac{\kappa}{\kappa + 1}} \mathbf{H}^{\mathrm{LoS}} + \sqrt{\frac{1}{\kappa + 1}} \mathbf{H}^{\mathrm{NLoS}} \right),
    \label{eq:channel_model}
\end{equation}
where $\beta$ represents the large-scale fading coefficient, and $\kappa = A \exp(B \theta)$ is the Rician factor (with environment-dependent constants $A$ and $B$), denoting the power ratio of the LoS component to the NLoS multipath components. This modeling captures the empirical observation that higher elevation angles typically lead to stronger LoS components.

\textbf{Large-Scale Fading:} The coefficient $\beta$ incorporates distance-dependent path loss and shadow fading, expressed in linear scale as $\beta = 10^{-(L + \xi)/10}$. Here, $\xi \sim \mathcal{N}(0, \sigma_{sh}^2)$ represents the log-normal shadowing, and $L = P_{\mathrm{LoS}}(z) \cdot L^{\mathrm{LoS}} + (1 - P_{\mathrm{LoS}}(z)) \cdot L^{\mathrm{NLoS}}$ is the path loss determined by the height-dependent LoS probability $P_{\mathrm{LoS}}(z)$.

\textbf{LoS Component:} The LoS component is determined by the array steering vectors and the Doppler shift, expressed as 
\begin{align}
    \mathbf{H}^{\mathrm{LoS}} = \mathbf{a}_{\mathrm{BS}}(\phi^{\mathrm{A}}, \theta^{\mathrm{A}}) \mathbf{a}_{\mathrm{UAV}}^H(\phi^{\mathrm{D}}, \theta^{\mathrm{D}}) e^{j 2\pi \nu^{\mathrm{LoS}} t},
\end{align}
where $\phi^{\mathrm{D}}$ and $\theta^{\mathrm{D}}$ represent the azimuth and elevation Angles of Departure (AoD) at the UAV, while $\phi^{\mathrm{A}}$ and $\theta^{\mathrm{A}}$ denote the Angles of Arrival (AoA) at the BS. The Doppler shift is $\nu^{\mathrm{LoS}} = \frac{1}{\lambda} \mathbf{v}^T \mathbf{r}$, where $\lambda$ is the carrier wavelength and $\mathbf{r}$ specifies the unit direction vector. 

\begin{figure*} [t]
	\centering 
		 \includegraphics[width=0.98\textwidth]{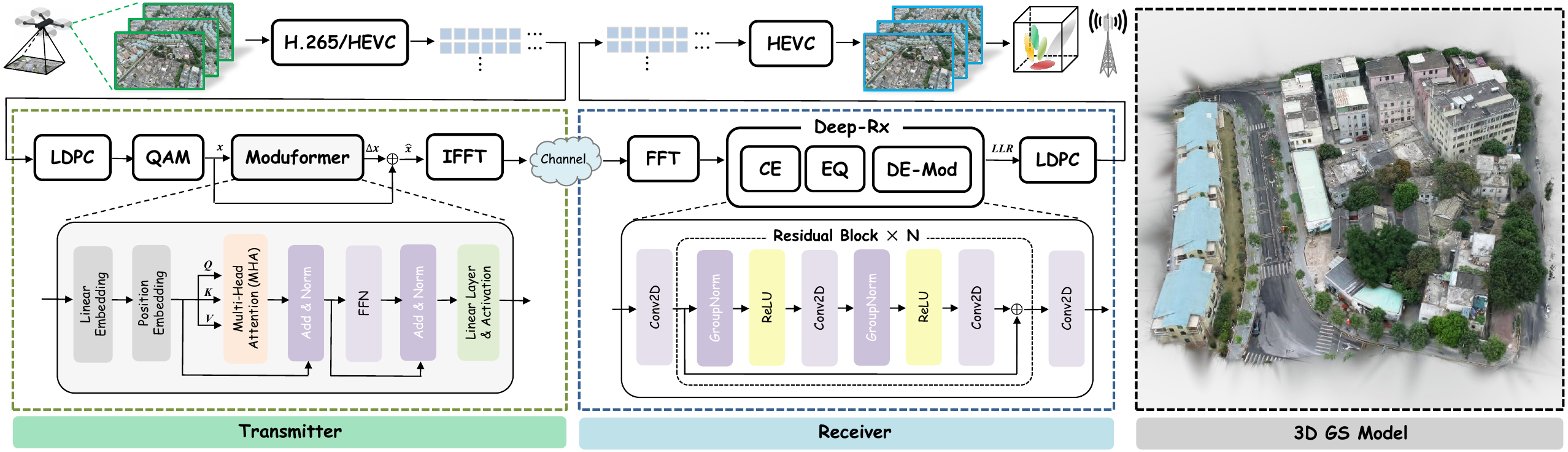}  
	
 	\caption{Architecture of the proposed task-oriented end-to-end transceiver. The transmitter utilizes a Transformer-based Moduformer for resilient constellation mapping, while the receiver employs a Deep-Rx network. Both modules are jointly optimized driven by the 3DGS rendering loss.}  \vspace{-0.2in}\label{Fig2}
\end{figure*}

Assuming $N_{\mathrm{BS}} = N_y N_z$, the array response vector for the BS UPA is given by:
\begin{align}
    \mathbf{a}_{\mathrm{BS}}(\phi, \theta) =& \frac{1}{\sqrt{N_{\mathrm{BS}}}} \big[ 1, \dots, e^{j \frac{2\pi d_a}{\lambda} (n_y \sin\phi \sin\theta + n_z \cos\theta)}, \dots \nonumber \\ 
    & \dots, e^{j \frac{2\pi d_a}{\lambda} ((N_y-1) \sin\phi \sin\theta + (N_z-1) \cos\theta)} \big]^T,
\end{align}
where $0 \le n_y < N_y$, $0 \le n_z < N_z$, and $d_a$ is the antenna spacing. Similarly, $\mathbf{a}_{\mathrm{UAV}}$ denotes the uniform linear array response vector for the UAV.

\textbf{NLoS Component:} The NLoS component models the scattering environment and consists of $K$ propagation clusters:
\begin{equation}
    \mathbf{H}^{\mathrm{NLoS}} = \frac{1}{\sqrt{K}} \sum_{k=1}^{K} g_{k} \mathbf{a}_{\mathrm{BS}}(\phi_{k}^{\mathrm{A}}, \theta_{k}^{\mathrm{A}}) \mathbf{a}_{\mathrm{UAV}}^H(\phi_{k}^{\mathrm{D}}, \theta_{k}^{\mathrm{D}}) e^{j 2\pi \nu_{k} t},
\end{equation}
where $g_{k} \sim \mathcal{CN}(0, 1)$ is the complex gain of the $k$-th path, and $\nu_{k}$ is the associated Doppler shift.

\subsection{Signal Transmission and Reception}
The UAV transmits a data stream containing the compressed aerial images to the BS. Let $\mathbf{s} \in \mathbb{C}^{N_{\mathrm{UAV}} \times 1}$ denote the transmitted signal vector. The received signal vector $\mathbf{y} \in \mathbb{C}^{N_{\mathrm{BS}} \times 1}$ at the BS is:
\begin{equation}
    \mathbf{y} = \mathbf{H} \mathbf{s} + \mathbf{n},
    \label{eq:received_signal}
\end{equation}
where $\mathbf{n} \sim \mathcal{CN}(\mathbf{0}, \sigma^2 \mathbf{I}_{N_{\mathrm{BS}}})$ denotes the additive white Gaussian noise (AWGN) vector. Assuming the BS employs a linear combining vector or a learning-based neural network decoder to process $\mathbf{y}$, the transmitted bitstreams can be efficiently recovered for downstream 3D scene reconstruction.

\section{Proposed Transceiver Design}
In general, conventional transceiver designs employ eigen zero-forcing (EZF) precoding~\cite{EZF} at the transmitter, while relying on least squares (LS) or weighted minimum mean square error (WMMSE) algorithms~\cite{wmmse} for channel estimation followed by EZF for equalization at the receiver. 

However, the performance of such modular processing architectures are limited, as they independently optimize each component and heavily rely on the acquisition of perfect channel state information (CSI). To bypass the need for explicit mathematical models and achieve better performance, the paradigm has gradually shifted towards E2E deep learning methods.

Recently, in many advanced deep learning-based transceiver designs, constellation point learning is widely adopted. Let $\mathcal{C} = \{c_1, c_2, \dots, c_M\}$ denote the set of constellation points for an $M$-ary modulation scheme. Existing methods treat these constellation coordinates $c_m \in \mathbb{C}$ directly as trainable parameters, optimizing their geometric distribution, which can be written as
\begin{align}
    \tilde{c} = \frac{c - \frac{1}{2^M} \sum_{i=1}^{2^M} c_i}{\sqrt{\frac{1}{2^M} \sum_{i=1}^{2^M} |c_i|^2 - \left| \frac{1}{2^M} \sum_{i=1}^{2^M} c_i \right|^2}}.
\end{align}
While this approach has primarily been validated in single-input multiple-output (SIMO) systems, it exhibits severe training instability and degraded performance when applied to MIMO systems under dynamic low-altitude UAV channels.

A primary reason for this bottleneck is the imbalance in the number of trainable parameters between the transmitter and the receiver. While the neural receiver (e.g., Deep-Rx) typically employs deep convolutional networks with millions of parameters to combat complex fading, the transmitter only possesses $2^M$ parameters (e.g., 32 parameters for 16-QAM). During the E2E joint training via backpropagation, this severe parameter asymmetry causes the transmitter's gradients to vanish or fluctuate wildly, making it incapable of learning robust constellation representations for harsh A2G channels.

\subsection{Transmitter Architecture and Moduformer}

To overcome this fundamental limitation, this paper proposes a Transformer-based modulation architecture, termed \textbf{Moduformer}, utilizing a residual learning strategy. As illustrated in the transmitter module, standard QAM mapping is first applied to generate an initial, coarse symbol sequence $\mathbf{x} = [x_1, \dots, x_N]^T \in \mathbb{C}^{N \times 1}$. Rather than directly learning the absolute constellation positions, the Moduformer takes $\mathbf{x}$ as input to dynamically generate a sequence of channel-resilient perturbations $\Delta \mathbf{x} \in \mathbb{C}^{N \times 1}$. The ultimately transmitted frequency-domain symbol vector $\hat{\mathbf{x}}$ is obtained via residual addition: 
\begin{align}
    \hat{\mathbf{x}} = \mathbf{x} + \Delta\mathbf{x}.
\end{align}

The internal structure of the Moduformer leverages the self-attention mechanism to capture long-range dependencies among the transmitted symbols. The input complex sequence $\mathbf{x}$ is first projected into a higher-dimensional real-valued latent space via linear embedding, and superimposed with positional embeddings to form the initial state $\mathbf{E}_0$. The features are then processed by a multi-head attention (MHA) module and a feed-forward network (FFN), both accompanied by residual connections and layer normalization (Add \& Norm):
\begin{align}
    \mathbf{E}_{\mathrm{attn}} &= \mathrm{LayerNorm}\big(\mathbf{E}_0 + \mathrm{MHA}(\mathbf{E}_0)\big), \\
    \mathbf{E}_{\mathrm{out}} &= \mathrm{LayerNorm}\big(\mathbf{E}_{\mathrm{attn}} + \mathrm{FFN}(\mathbf{E}_{\mathrm{attn}})\big).
\end{align}
Finally, a linear layer with activation maps the latent representation $\mathbf{E}_{\mathrm{out}}$ back to the complex domain to yield the learned perturbation $\Delta \mathbf{x}$.

\begin{algorithm}[t]
\caption{E2E Training Algorithm}
\label{alg:e2e_training}
\begin{algorithmic}[1]
\REQUIRE Aerial images $\mathcal{I}$, camera poses $\mathcal{P}$, UAV channel $\mathbf{H}$, learning rate $\eta$.
\ENSURE Optimal network parameters $\Theta_{\text{Tx}}^*$ and $\Theta_{\text{Rx}}^*$.
\FOR{epoch $e = 1$ to $E_{\text{max}}$}
    \FOR{each mini-batch $(\mathbf{I}, \mathbf{p}) \in (\mathcal{I}, \mathcal{P})$}
        \STATE \textbf{Tx:} Encode $\mathbf{I}$ to bitstreams $\mathbf{b}$ and initial QAM symbols $\mathbf{x}$.
        \STATE \textbf{Tx:} Compute $\hat{\mathbf{x}} = \mathbf{x} + \text{Moduformer}(\mathbf{x}; \Theta_{\text{Tx}})$ and multiplex sparse pilots to form transmitted signal $\mathbf{s}$.
        \STATE \textbf{Channel:} Receive signal $\mathbf{y} = \mathbf{H}\mathbf{s} + \mathbf{n}$.
        \STATE \textbf{Rx:} Process $\mathbf{y}$ via $\text{Deep-Rx}(\mathbf{y}; \Theta_{\text{Rx}})$ to recover bits $\hat{\mathbf{b}}$ and images $\hat{\mathbf{I}}$.
        \STATE \textbf{3D-GS:} Render novel views $I_{\text{render}}$ using recovered images $\hat{\mathbf{I}}$ and poses $\mathbf{p}$.
        \STATE \textbf{Loss:} $\mathcal{L}_{\mathrm{total}} = \mathcal{L}_{\mathrm{BCE}}(\mathbf{b}, \hat{\mathbf{b}}) + \psi \big( (1-\alpha)\mathcal{L}_1(\mathbf{I}, I_{\text{render}}) + \alpha \mathcal{L}_{\mathrm{SSIM}}(\mathbf{I}, I_{\text{render}}) \big)$.
        \STATE \textbf{Update:} $\Theta \leftarrow \Theta - \eta \nabla_{\Theta} \mathcal{L}_{\mathrm{total}}$, where $\Theta \in \{\Theta_{\text{Tx}}, \Theta_{\text{Rx}}\}$.
    \ENDFOR
\ENDFOR
\end{algorithmic}
\end{algorithm}

The proposed Moduformer architecture fundamentally resolves the parameter imbalance by equipping the transmitter with sufficient learning capacity. Crucially, because the self-attention mechanism embeds rich, long-range structural correlations directly into the data payload, the system's reliance on traditional, dense pilot sequences for channel estimation is significantly reduced. This naturally facilitates a highly efficient, sparse-pilot transmission scheme that maximizes the effective throughput for aerial imagery. Furthermore, the perturbation learning strategy ensures that the training starts from a solid baseline (standard QAM), requiring the network only to learn the necessary geometric distortions to combat specific channel fading. Ultimately, this synergy drastically accelerates convergence, improves E2E training stability, and delivers superior robustness in low-altitude environments.

\subsection{Receiver Architecture and Loss Function Design}
ResNet-based Deep-Rx architecture has been widely established as highly effective solutions for E2E transceiver designs~\cite{Nvidia}. By replacing isolated signal processing blocks with a unified convolutional network, this architecture inherently mitigate the cascading errors typical of modular designs, adeptly executing joint channel estimation, equalization, and demodulation through robust non-linear mappings. 

However, when applied to downstream large-scale 3D scene reconstruction, conventional receiver designs are fundamentally limited by their strict optimization for isolated physical layer metrics (e.g., block error rate) or simple image-level losses (e.g., mean squared error). These metrics often fail to capture the complex geometric consistency and high-frequency textural details strictly required for robust 3DGS.

To bridge this gap between the communication link and the downstream reconstruction task, this paper proposes a task-driven E2E optimization strategy. Rather than isolating these stages, the proposed method seamlessly integrate the 3DGS rendering performance into the transceiver's training process. Our approach directly employ the 3D scene rendering loss to guide the parameter updates of both the transmitter's Moduformer and the receiver's Deep-Rx. The E2E task-driven loss function can be formulated as:
\begin{equation}
    \mathcal{L}_{\mathrm{total}} = \mathcal{L}_{\mathrm{BCE}} + \psi \Big( (1-\alpha)\mathcal{L}_1 + \alpha \mathcal{L}_{\mathrm{SSIM}} \Big),
    \label{eq:e2e_loss}
\end{equation}
where $\mathcal{L}_{\mathrm{BCE}}$ denotes the binary cross-entropy loss for bit recovery at the receiver. For the novel view synthesis task, $\mathcal{L}_1$ represents the $\mathcal{L}_1$ photometric loss measuring the pixel-wise absolute difference between the rendered and ground-truth images, while $\mathcal{L}_{\mathrm{SSIM}}$ incorporates the Structural Similarity Index (SSIM) to preserve high frequency structural details and perceptual quality.
In the second term, $\psi$ serves as a task-weighting hyperparameter to align the scale of the gradients between the two domains, and $\alpha$ strictly balances the structural perception of the SSIM index with the pixel level fidelity. The overall architecture of the transceiver design is shown in~Fig.~\ref{Fig2}.

\begin{figure} [t]
	\centering 
		 \includegraphics[width=0.42\textwidth]{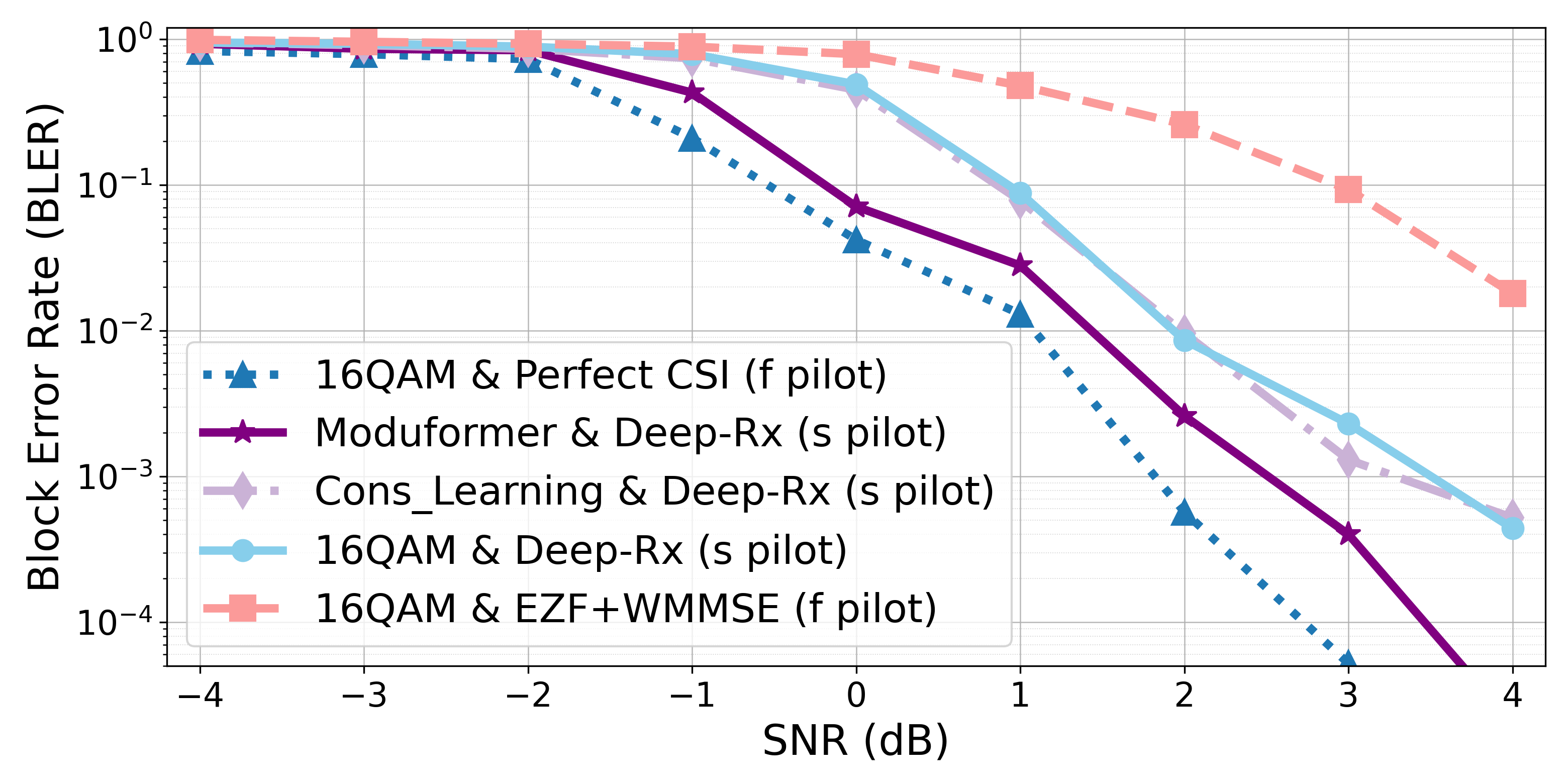} 
  	\caption{BLER comparison of various transmission schemes versus SNR.}\vspace{-0.2in}  \label{Fig3}
\end{figure}

\subsection{E2E Training Scheme}
The detailed E2E training procedure is summarized in Algorithm \ref{alg:e2e_training}. During the forward pass, the source images $\mathcal{I}$ are processed by the Moduformer at the transmitter and transmitted over the simulated dynamic UAV channel. Upon reception, the Deep-Rx module jointly recovers the bitstreams and images, which are subsequently rendered by the 3DGS framework to evaluate the rendering loss. The network parameters of both the transmitter and receiver, denoted jointly as $\Theta \triangleq \{\Theta_{\text{Tx}}, \Theta_{\text{Rx}}\}$, are concurrently updated using a stochastic gradient descent-based optimizer with a learning rate $\eta$:$$\Theta^{*} = \arg\min_{\Theta} \mathbb{E}_{(\mathbf{I}, \mathbf{p}) \in (\mathcal{I}, \mathcal{P})} \big[ \mathcal{L}_{\mathrm{total}}(\Theta) \big],$$where the expectation is taken over the mini-batches of images and their corresponding camera poses. This unified parameter update mechanism couples the physical layer representations with the geometric priors required for 3D reconstruction.

\begin{figure} [t]
	\centering 
		 \includegraphics[width=0.42\textwidth]{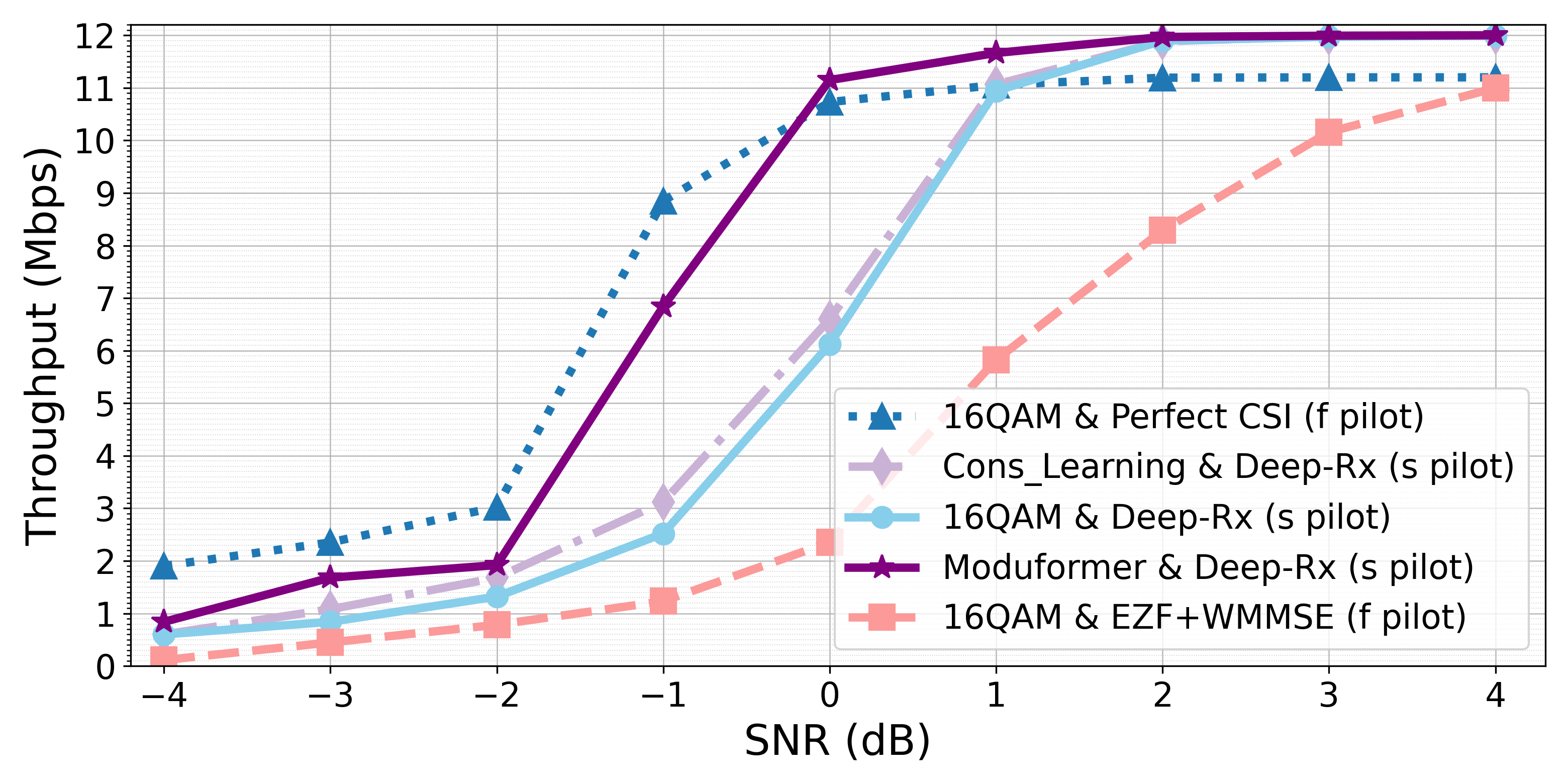} 
  	\caption{Effective throughput comparison across varying SNRs.}\vspace{-0.2in}  \label{Fig4}
\end{figure}

\section{Experiments}
We consider a wireless system where an UAV equipped with $N_{UAV} = 4$ antennas transmits signals to a BS featuring a massive MIMO array of $N_{BS} = 64$ antennas. The system operates in a single-stream configuration to focus on the reliability of the underlying link. For error correction, a 5G NR-compliant low-density parity-check (LDPC) scheme is employed with a code rate of $r = 0.5$. The waveform is based on orthogonal frequency division multiplexing (OFDM), characterized by $N_c = 72$ subcarriers and $N_s = 14$ OFDM symbols per slot, forming a resource grid of $72 \times 14$ resource elements (REs).

Full Pilot ($f\text{-pilot}$): Represents the conventional pilot-assisted baseline. It reserves two complete OFDM symbols per slot for pilot transmission, providing dense reference signals to the receiver. Sparse Pilot ($s\text{-pilot}$): Specifically, $s\text{-pilot}$ only occupies a fraction of the resource grid (e.g., a comb-type pattern in a single symbol), reducing the pilot overhead by 90\% compared to $f\text{-pilot}$.

\subsection{Datasets}
To evaluate our method in large-scale, real-world scenarios, we utilize extensive UAV imagery from the GauU-Scene~\cite{xiong2024gauuscenev2assessingreliability} and Mega-NeRF~\cite{Turki_2022_CVPR} datasets.

\subsubsection{Training Set}
For robust generalization, our training set encompasses 20,000 high-resolution UAV images from two main sources, enabling the model to learn comprehensive geometric and appearance priors:
\begin{itemize}
    \item \textbf{GauU-Scene V2:} Four sub-datasets (SZIIT, CUHK-SZ Lower/Upper Campus, and LFLS) covering diverse academic and urban environments.
    \item \textbf{Mega-NeRF (Mill 19):} The Building and Rubble scenes, which introduce industrial and debris-filled textures.
\end{itemize}

\begin{figure} [t]
	\centering 
		 \includegraphics[width=0.46\textwidth]{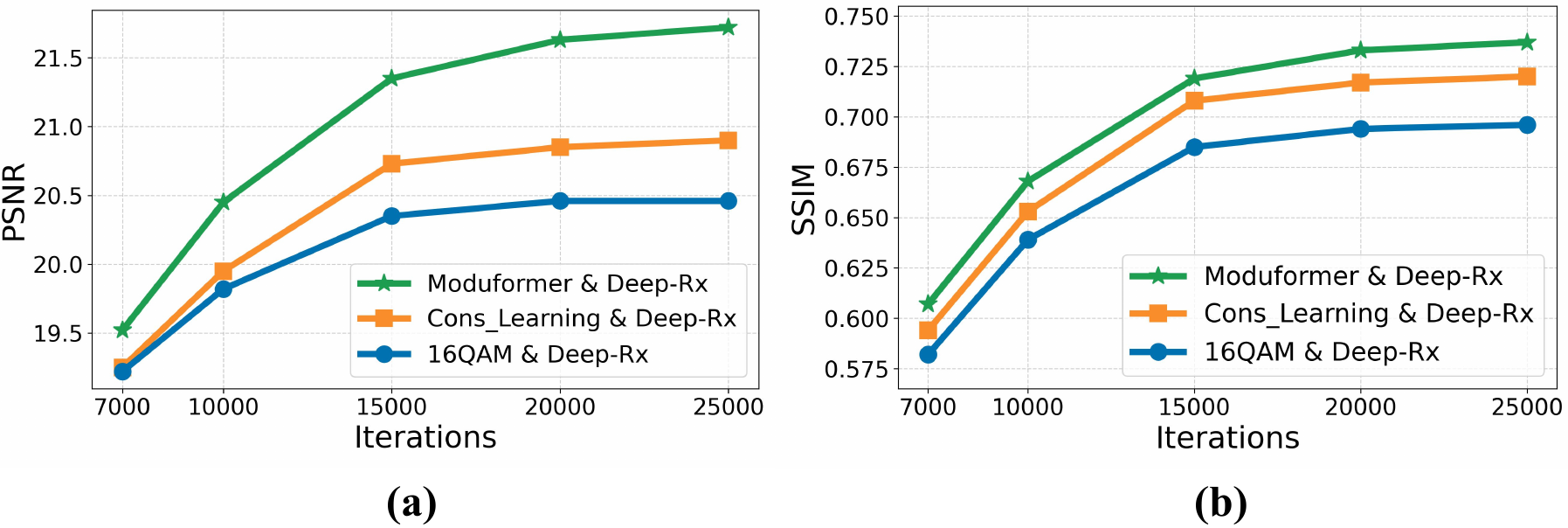} 
  	\caption{Quantitative evaluation of novel view synthesis quality. (a) PSNR versus training iterations.  (b) SSIM versus training iterations.}\vspace{-0.16in}  \label{Fig5}
\end{figure}

\begin{figure} 
	\centering 
		 \includegraphics[width=0.49\textwidth]{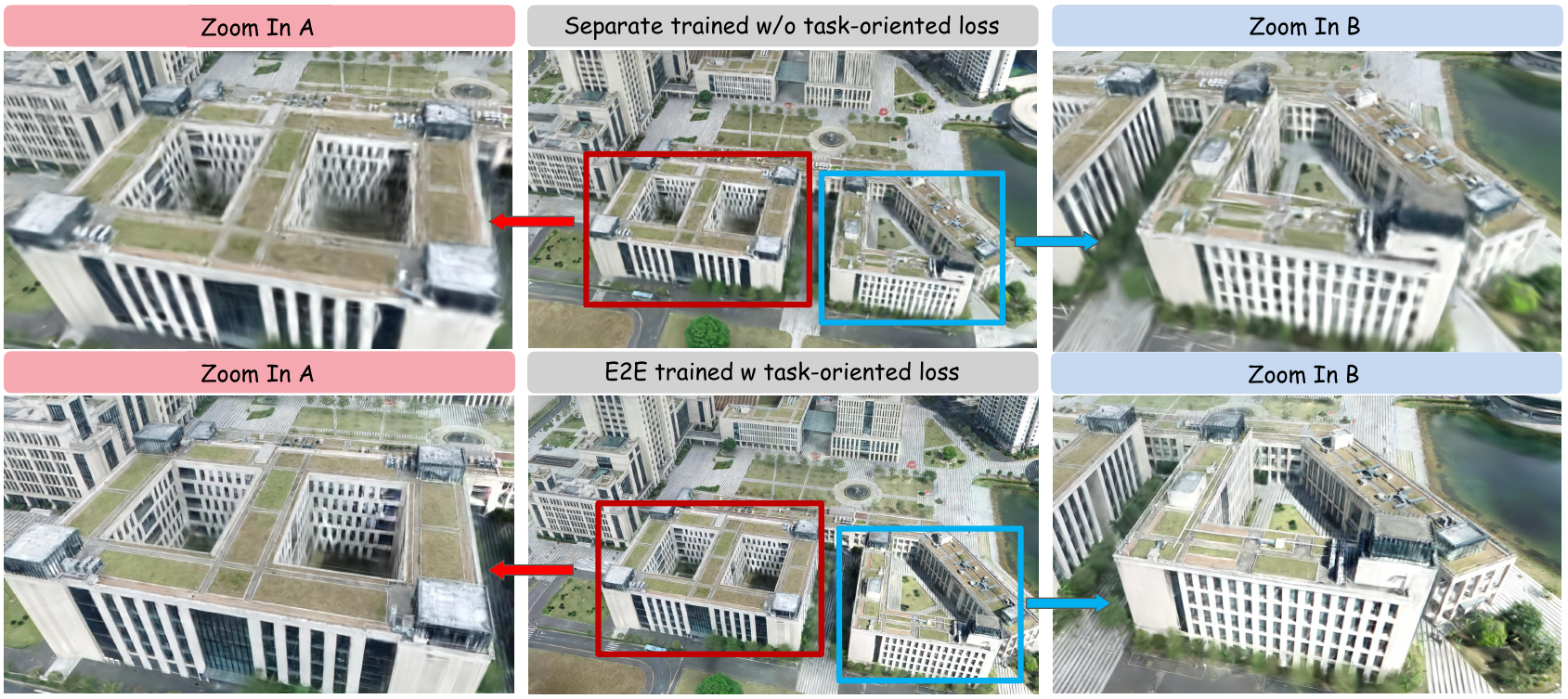} 
  	\caption{Qualitative visualization with and without the task-oriented loss.}\vspace{-0.25in}  \label{Fig6}
\end{figure}

\subsubsection{Testing Set}
For evaluation, we specifically employ the SMBU and HAV sub-datasets from GauU-Scene. These two datasets provide highly complex urban topology and campus scene, serving as rigorous benchmarks to test the model's reconstruction fidelity and overall performance limits.

\subsection{Performance Comparision}
We first evaluate transmission performance using BLER and throughput to quantify link reliability and effective data rate. Our proposed Moduformer \& Deep-Rx (s pilot) is compared against four schemes. 16QAM \& Perfect CSI (f pilot) serves as the theoretical upper bound. Practical baselines include the conventional 16QAM \& EZF+WMMSE (f pilot), alongside two sparse-pilot learning variants: Cons-Learning \& Deep-Rx~\cite{aibo} and 16QAM \& Deep-Rx. 

As shown in~Fig.~\ref{Fig3} and~Fig.~\ref{Fig4}, evaluating across an SNR range of -4 to 4 dB, our method significantly outperforms both conventional and learning-based baselines in BLER. Compared to the constellation learning approach, the Moduformer better balances parameter scale and training stability. This ensures robust symbol reconstruction that closely approaches the perfect CSI bound, despite utilizing minimal pilot information. Furthermore, our design consistently achieves the highest throughput among all practical schemes. This advantage is directly attributed to the sparse pilot strategy, which significantly minimizes reference overhead and maximizes the spectral efficiency of the end-to-end system.

\begin{figure} [t]
	\centering 
		 \includegraphics[width=0.46\textwidth]{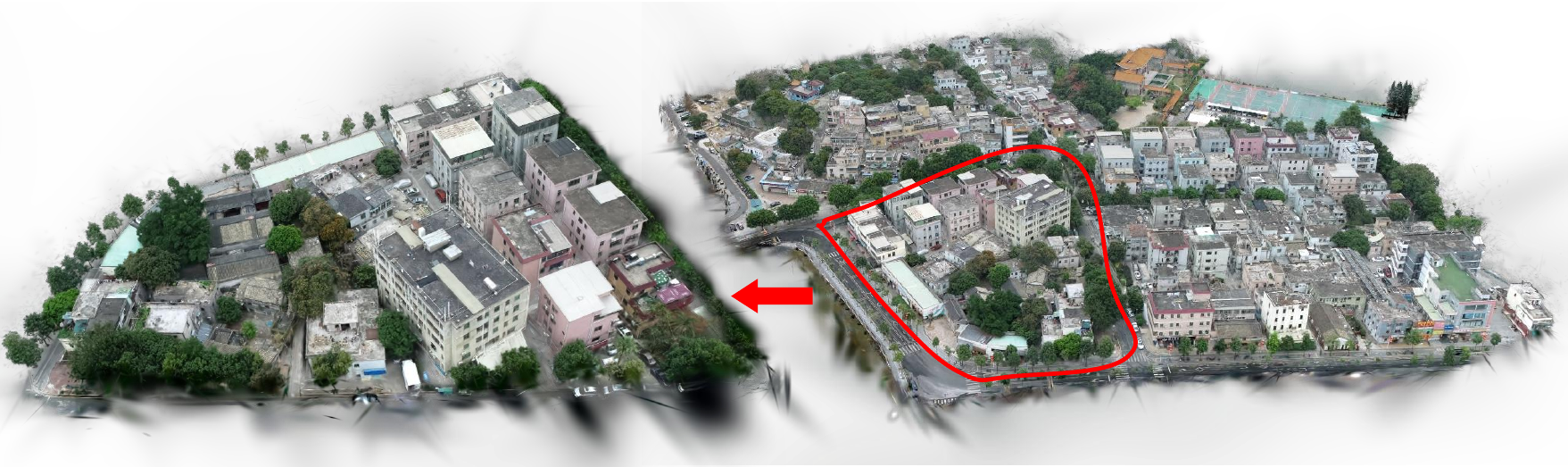} 
  	\caption{Qualitative visualization of the reconstructed 3D scene (HAV).}\vspace{-0.2in}  \label{Fig7}
\end{figure}

Finally, we evaluate the novel view synthesis quality of the reconstructed 3DGS models. As shown in~Fig.~\ref{Fig5}, our proposed method consistently yields higher peak signal-to-noise ratio (PSNR) and SSIM across all training iterations compared to both the constellation learning and 16QAM baselines. This improvement highlights the advantage of Moduformer, as the Transformer-based transmitter effectively captures global spatial dependencies and balances the learning capacity between the transmitter and receiver. In~Fig.~\ref{Fig6} it can be observed that, utilizing the task-oriented loss, integrating 3DGS into the training process prioritizes critical geometric features during modulation, and this rendering-aware loss ensures robust scene reconstruction compared to the separate training scheme. The qualitative superiority of our method is further visually corroborated by the rendered 3D scene in~Fig.~\ref{Fig7}.

\section{Conclusion}
This paper proposed a task-oriented end-to-end transceiver with a sparse pilot scheme for efficient aerial image transmission and robust 3D scene reconstruction. To reduce pilot overhead, we introduced a Transformer-based Moduformer for channel-resilient modulation and a Deep-Rx network for joint signal recovery. By integrating 3D Gaussian Splatting (3DGS) directly into the training process, the communication modules are jointly optimized via downstream rendering loss. Evaluations on real-world datasets demonstrate that the proposed Moduformer improves PSNR by 6.2\% and SSIM by 5.3\%, while maintaining a strictly lower BLER and boosting throughput by approximately 78\% at 0 dB SNR.

\clearpage
\balance
\bibliographystyle{IEEEtran}
\bibliography{ref}

\end{document}